\documentclass{article}

\usepackage{arxiv}

\usepackage[utf8]{inputenc} 
\usepackage[T1]{fontenc}    
\usepackage{hyperref}       
\usepackage{url}            
\usepackage{booktabs}       
\usepackage{amsfonts}       
\usepackage{nicefrac}       
\usepackage{microtype}      
\usepackage{cite}
\usepackage{amsmath,amssymb,amsfonts}
\usepackage{algorithmic}
\usepackage{graphicx}
\usepackage{textcomp}

\title{Contextual Skipgram: Training Word Representation Using Context Information}
\author{
Dongjae Kim \\
School of Electrical Engineering\\ Korea University\\
Anam Dong, Seoul, South Korea\\
\And
Jong-Kook Kim \\
School of Electrical Engineering\\ Korea University\\
Anam Dong, Seoul, South Korea\\
}

\begin{document}
\maketitle

\begin{abstract}
The skip-gram (SG) model learns word representation by predicting the words surrounding a center word from unstructured text data. However, not all words in the context window contribute to the meaning of the center word. For example, less relevant words could be in the context window, hindering the SG model from learning a better quality representation. In this paper, we propose an enhanced version of the SG that leverages context information to produce word representation. The proposed model, Contextual Skip-gram, is designed to predict contextual words with both the center words and the context information. This simple idea helps to reduce the impact of irrelevant words on the training process, thus enhancing the final performance.
\end{abstract}

\keywords{ machine learning \and word embedding \and skip-gram \and language model}

\section{Introduction}
Distributed representations of words have been an essential approach to achieving good performance in natural language processing (NLP) tasks. Most deep learning NLP models use pre-trained word embeddings for successful training. Earlier word embedding models are trained based on Neural Network Language Model (NNLM) which involves dense matrix multiplications \cite{Bengio:2003:NPL:944919.944966, Collobert:2008:UAN:1390156.1390177}. As a result, they need long training times when training large corpora.

Two popular word representation models, the skip-gram (SG) and continuous bag-of-words (CBOW), were proposed in 2013 \cite{41224, Mikolov:2013:DRW:2999792.2999959}. The main idea behind these models is that words that are similar to each other are likely to share a similar co-occurrence of nearby words. Predicting the surrounding words with the center word or vice versa is how the SG and CBOW models train word representation. This training process is done by moving a sliding window through the corpus. As a result, the $|V| * d$ embedding matrix is trained, where $V$ and $|V|$ refer to the vocabulary and the number of different vocabularies in the given corpus respectively, and $d$ is a hyperparameter defining the dimension of each word vector. Because Word2Vec models do not have non-linear hidden layers, they can process a large corpus much faster than the earlier NNLM based models.

The SG architecture learns word embeddings by predicting contextual words given a center word, while the CBOW architecture learns by predicting center word given contextual words. Because the CBOW compacts nearby word vectors into a single average vector, CBOW executes its task faster. On the other hand, the SG has more chances of learning with the same size of corpus compared to the CBOW, because all possible contextual words and center word pairs are used for learning. As a result, Skip-gram tends to work better with the a smaller corpus. For a large $|V|$, using the softmax function requires a great deal of computation. Thus negative sampling, which is simplified variant of Noise Contrastive Estimation (NCE) \cite{Gutmann:2012:NEU:2503308.2188396}, is preferred for training a large corpus.

Due to their huge success in many NLP tasks, there have been various studies on improving the performance of the SG model and leveraging external linguistic resources such as semantic lexicons is one of them. The information from the resources are incorporated to refine objective function \cite{yu-dredze-2014-improving, kiela-etal-2015-specializing} or utilized in a retrofitting scheme \cite{kiela-etal-2015-specializing, faruqui-etal-2015-retrofitting}. Though these approaches improve the semantic quality of the SG model, they require reliable external linguistic resources which are hard to obtain and produce.

To train better word vectors with only a given corpus, \cite{ling-etal-2015-two} proposed leveraging word order information in the local context window. The structured skip-gram (SSG) and continuous window (CWin) models increased output embedding size proportional to the context window size. \cite{song-etal-2018-directional} introduced directional skip-gram (DSG) and simplified structured skip-gram(SSSG) models to train with direction information. These approaches showed improvement in some word similarity and Part-of-Speech (POS) tagging tasks.

Fasttext model proposed representing each word as a bag of character n-grams to better utilize the morphological information of the words \cite{bojanowski-etal-2017-enriching}. Even if specific words are rarely seen in the corpus, their subwords or subfeatures can be trained during the training of other full words. As a result, this model provides a better rare word embedding quality. Moreover, word embeddings that are not seen in the training corpus can be estimated with the subword information.

The difference between previous methods and our proposed method, the Contextual Skip-Gram (CSG) model, is that our scheme uses a context vector built from the local context window. As shown in the results, the CSG model was able to achieve overall good performance for similarity tasks for both small and large corpora compared to conventional models. Section 2 describes our method, while the experiments and results are depicted in Section 3. Section 4 summarizes the research.

\section{Contextual Skip-Gram}
The SG model learns word embeddings by predicting nearby words given the center word. Thus, the training objective of the SG model is to maximize the overall log probability: $$\mathcal{L}_{SG} = {\frac{1}{|V|}} \sum_{t=1}^{|V|} \sum_{0 < |i| \leq c} \log p(w_{t+i} | w_t)$$ where $w_t$ and $w_{t+i}$ refer to the center word and nearby word to predict. Given sampled negative word set $V^-$, its negative sampling objective is defined: $$\log\sigma((v_{w_t}\top v'_{w_{t + i}})) + \sum_{w_j \in V^-}\log\sigma((v_{w_t}\top v'_{w_j}))$$ where $v$ and $v'$ are the input and output vector representations of corresponding word.

While all the context words participate in building the center word vector, they do not always make equal contribution.
\begin{quote}
\textit{Water becomes solid ice when it is cold enough.}
\end{quote}
In the example sentence above, the following pairs can be made when the window size is five and the center word is ice. 
\begin{quote}
\textit{(ice, water) (ice, becomes), (ice, solid), (ice, when), (ice, it),  (ice, cold), (ice, enough)}
\end{quote}
In human sense, \textit{"water"} and \textit{"cold"} contribute to the meaning of the center word \textit{"ice"} more than other context words such as \textit{"when"} and \textit{"becomes"}. In the SG manner, given center word \textit{"ice"}, \textit{"water"} and \textit{"cold"} should be more predictable. Though \textit{"when"} and \textit{"becomes"} could contribute syntactically to the meaning of \textit{"ice"}, their co-occurrence is less convincing semantically. We call words less relevant to the center word as weak co-occurrences and they could potentially disturb training due to the nature of the SG. The weak co-occurrences could be frequently used words such as articles, typo or etc. Generally, large training corpus relieves this issue. As long as training corpus is large enough, \textit{water} and \textit{ice} are likely to share similar context word co-occurrence and weak co-occurences \textit{"when"} is likely to be used with many other words. As a result, the SG model could learn descent word representations without human effort such as annotation.

However, as training progress, this could still cause degraded performance. At the initial epoch, both relevant and less relevant nearby words would have low probabilities given a center word, because they are randomly initialized. However, as the SG model trains through the corpus, less relevant pairs are likely to have lower probabilities than relevant words. Consequently, the SG model would increase the probabilities of weak co-occurrences rather than more relevant words to increase the overall prediction probability. This issue could hinder word embeddings from acquiring higher quality. The experiment in Section \ref{prediction analysis} addresses this issue in detail.

This problem is mainly due to SG model's nature. it trains word embeddings assuming a direct relationship between the center word and the surrounding word. To alleviate this issue, the CSG model predicts nearby words with context information to introduce indirect relationships. Our objective is to maximize following loss: $$\mathcal{L}_{CSG} = {1 \over |V|} \sum_{t=1}^{|V|} \sum_{0 < |i| \leq c} \log p(w_{t+i} | w_t, w_{con})$$ where $w_t$, $w_{t+i}$ and $w_{con}$ refer to the center word, nearby word to predict and the surrounding words as context information. $c$ is a hyperparameter that defines the context window size. Our negative sampling loss is defined as: $$ s(v'_{w_{t+i}}, v_{w_t}, v_{con}) + \sum_{w_j \in V^-} s(-v'_{w_j}, v_{w_t}, v_{con}) $$ where $v$ and $v'$ denote the input and output embedding of corresponding words and $V^-$ denotes negative samples. $v_{con}$ and $s$ are further described in following Sections \ref{context fucntion} and \ref{weighted fusion}.

\subsection{Context Functions} \label{context fucntion}
To make probability calculation between context information and nearby words easy, we first aggregate the context information into context embedding $v_{con}$, which has the dimension $d$ via the context function. The most simple way to make context embedding is to average input embeddings of surrounding words. However, we use a summing function to make vector updates simple and fast. The Section \ref{update rule} describes detailed reason.

\subsection{Weighted Fusion} \label{weighted fusion}
The prediction probabilities based on center word and context embedding should be combined to produce the final probability. The CSG model has two different fusion strategies to combine prediction probabilities.

The \textbf{Early Fusion (EF)} method executes the element-wise weighted sum of two vectors first and then calculates the log probability with fused vector and output embedding of nearby words $$ s_{EF} = \log\sigma((\gamma v_{con}  + (1 - \gamma) v_{w_t})\top v'_{w_{t+i}})$$ Function $\sigma$ in the above equation denotes the sigmoid function, and $\gamma$ is the fusion weight, which is a hyperparameter ranged $0 \leq \gamma \leq 1$. The early fusion is simple, but does not guarantee ratio due to the difference of vector magnitude between the context vector and the center word vector.

The \textbf{Late Fusion(LF)} method calculates dot products first, then performs a weighted summation. $$s_{LF} = \gamma\sigma(v_{con}\top \cdot v'_{w_{t+i}}) + (1 - \gamma)\sigma(v_{w_t}\top \cdot v'_{w_{t+i}})$$

The fusion weight $\gamma$ decides where to focus on during training. If $\gamma$ is high, predictions are more dependent on the context vector, while the center word vector controls minor adjustments. The value of $\gamma$ can be static or dynamic during training. In experiments, we used several fixed weights and linear weight scheme defined: $$\gamma_{linear, 0 \to 1} = {epoch_{current} -1 \over epoch_{total} -1 }$$ $$\gamma_{ran} = random_{uniform}(0, 1) $$ where $epoch_{current}$ denotes the current epoch count during training and $epoch_{total}$ is a hyperparameter defining the number of total epochs to learn.

\subsection{Update Rule} \label{update rule}
Given $\alpha$ as learning rate, the vector update in the SG model can be approximated as following equation:
$$ g = label - \sigma(v_{w_t}\top v'_{w_{k}}) $$
$$ v_{w_t} \mathrel{+}= \alpha \cdot g \cdot v_{w_{k}}' $$
$$ v_{w_{k}}' \mathrel{+}= \alpha \cdot g \cdot v_{w_t} $$
where $label$ is one for $w_k$ is positive sample and zero for negative samples. After gradient $g$ is calculated, vectors are updated according to the gradient and learning rate. The CSG model predict words with context representation $v_{con}$. Hence, the vector updates should happen to the surrounding words according to the loss. However, additional computation and word vector updates proportional to the window size could serious harm the training speed of the CSG model.

The CSG model with averaging function predicts nearby word with the center word and the context representation vector and they are weighted with $\gamma$. Given sentence $ S = \{ ..., w_{t-2}, w_{t-1}, w_{t}, w_{t+1}, w_{t+2}, ... \} $, the input vector representation of $v_{w_t}$ is updated when it is center word and when it is surrounding word. When $w_{t}$ is a center word, the weight of $w_{t}$ is $1 - \gamma$ according to weighted fusion. Otherwise, the weight of $w_{t}$ is ${ \gamma \over 2c }$ because context vector is an average of nearby words. As a result, the vector $v_{w_t}$ is updated as the following equations:
$$ v_{w_t} \mathrel{+}= (1 - \gamma) \cdot \alpha \cdot \sum_{0 < |i| \leq c} g_{t, t+i} \cdot v_{w_{t+i}}' $$
$$ v_{w_t} \mathrel{+}= \sum_{m = t - c }^{t + c} { \gamma \over 2c } \cdot \alpha \cdot \sum_{0 < |i| \leq c} g_{t, m+i} \cdot v_{w_{m+i}}' $$
$$ {t - c \leq m_t \leq t + c},  {t - 2c \leq m + i \leq t + 2c} $$
where $c$ is not zero. Since the position of $w_t$ and $w_m$ is close, they shares at least 50\% of surrounding words.

If we assume they share same surrounding words, equations above can be simplified as below: 
$$ v_{w_t} \mathrel{+}= \alpha \underset{\text{equals 1}}{\underline{((1 - \gamma) + \sum^{2c} { \gamma \over 2c })}} \cdot (\sum_{0 < |i| \leq c} g_{t, t+i} \cdot v_{w_{t+i}}') $$
As a result, we could approximate the vector update of the CSG as following algorithm. 
$$ g = label - s(v'_{w_{t+i}}, v_{w_t}, v_{con}) $$
$$ v_{w_t} \mathrel{+}= \alpha \cdot g \cdot v_{w_{t+i}}' $$
$$ v_{w_{t+i}}' \mathrel{+}= \alpha \cdot g \cdot v_{w_t}' $$
To preserve size of the magnitude of gradient, we use summing function with this update algorithm.

\section{Experiments} \label{experiments}
\subsection{Baseline Models}
In our experiments, we compared the CSG model to the CBOW and SG models from \cite{41224}, the DSG and SSSG from \cite{song-etal-2018-directional} and the FastText model from \cite{bojanowski-etal-2017-enriching}. For our test, we trained embeddings to be 200-dimensional. Context window size, negative sample size, and total iteration count were set to five. The starting learning rate was given as 0.025. Words that appeared less than five times were not used in training. In the case of FastText, n-grams ranging from 3 to 6 characters were used.

\subsection{CSG Parameters}
To explore the properties of the CSG parameters, we tested various combinations, increasing $\gamma$ by 0.25. The notation $CSG_{EF,0.25}$ represents the CSG trained with EF as fusion method and 0.25 as $\gamma$. Models trained with the $\gamma_{linear, 0 \to 1}$ scheme are denoted as $CSG_{EF or LF, 0 \to 1}$. CSG with $\gamma=0$ for EF and LF are omitted because they are equivalent to the SG. Similarly, CSG with $\gamma=1$ and LF is also discarded due to its equivalence to $\gamma=1$ and EF.

\subsection{Corpora}
We prepared two different corpora for training word embeddings. A large corpus was extracted from the latest Wikipedia dump and pre-processed with the following steps. We first lowercased the extracted text and split the text into sentences. Sentences that had less than ten tokens were filtered to remove partial sentences created during the splitting process. The large corpus was composed of about 4.14 billion tokens. A small corpus, a subset of the large corpus, was created by 1\% random sampling of sentences. The small corpus was composed of 39.3 million tokens.

\subsection{Prediction Analysis} \label{prediction analysis}
On this experiment, we manually analyzed how well the SG and CSG model predict surrouding word. To check prediction performance, we logged $ \sigma(v_{w_t}\top v'_{w_{k}}) $ for the SG and $ s_{EF}(v'_{w_{t+i}}, v_{w_t}, v_{con}) $ for the CSG with $\gamma$ = 0.5, when the center word is \textit{"ice"}. The Table \ref{table:prediction} shows the average logged values for epoch one and five. Weak co-occurences, such as \textit{"an", "for"}, tend to show poor predictions on the SG model and it become worse at epoch five. On the other hand, the CSG model tend to predict those words better as expected. For relevant words, the CSG showed slight decline in prediction due to the indirect relationship.

\begin{table}[!ht]
\begin{center}
\begin{tabular}{ r|r r| r r }
 \hline
 & \multicolumn{2}{c}{SG} & \multicolumn{2}{c}{CSG} \\
 & 1 & 5 & 1 & 5 \\
 \hline
 an & 36.65 & 35.75 & 54.30 & 69.24 \\
 \hline
 for & 34.22 & 24.87 & 53.19 & 46.20 \\
 \hline
 in & 26.38 & 18.60 & 47.76 & 52.85 \\
 \hline
 who & 35.37 & 33.85 & 64.99 & 76.74 \\
 \hline
 \hline
 hockey & 90.82 & 98.84 & 88.95 & 98.06 \\
 \hline
 water & 72.21 & 70.68 & 66.08 & 64.80 \\
 \hline
 winter & 70.47 & 87.22 & 73.72 & 86.91 \\
 \hline
 cream & 73.81 & 96.74 & 69.95 & 95.15 \\
\end{tabular}
\end{center}
\caption{Nearby word prediction performance experiment. All values are multiplied by 100}
\label{table:prediction}
\end{table}

\subsection{Word Similarity Evaluation}
To compare the performance of the embeddings, we performed a word similarity evaluation with Simlex-999 \cite{DBLP:journals/corr/HillRK14}, WordSim-353 \cite{Finkelstein:2001:PSC:371920.372094} and MEN-3k \cite{Bruni:2014:MDS:2655713.2655714} datasets. Similarity scores were measured by calculating the cosine similarity between two normalized word vectors for all pairs in each dataset. The Spearman's rank correlation coefficient between obtained scores and human judged scores was calculated.

\begin{table}[!ht]
\begin{center}
\begin{tabular}{ |r|r|r|r| }
 \hline
 & Sim-999 & WS-353 & MEN-3k\\
 \hline
 CBOW & 27.52 & 60.98 & 56.52 \\
 \hline
 SG & 33.57 & 66.15 & 63.36 \\
 \hline
 DSG & 33.04 & 65.48 & 63.09 \\
 \hline
 SSSG & 32.01 & 63.90 & 60.16 \\
 \hline
 FastText & 32.17 & 64.65 & \textbf{65.17} \\
 \hline
 \hline
 $CSG_{EF,0.25}$ & 33.50 & 67.24 & 63.94 \\
 \hline
 $CSG_{LF,0.25}$ & 33.47 & 66.90 & 63.90 \\
 \hline
 $CSG_{EF,0.5}$ & 33.42 & 67.58 & 63.75 \\
 \hline
 $CSG_{LF,0.5}$ & 33.48 & 67.75 & 64.02 \\
 \hline
 $CSG_{EF,0.75}$ & 33.27 & 67.42 & 63.42 \\
 \hline
 $CSG_{LF,0.75}$ & 33.26 & 67.57 & 63.63 \\
 \hline
 $CSG_{EF,1}$ & 32.67 & 66.16 & 62.95 \\
 \hline
 $CSG_{EF,0 \to 1}$ & 34.70 & \textbf{68.29} & 63.39 \\
 \hline
 $CSG_{LF,0 \to 1}$ & \textbf{34.95} & 67.84 & 62.46 \\
 \hline
 $CSG_{EF,ran}$ & 32.86 & 67.39 & 63.44 \\
 \hline
 $CSG_{LF,ran}$ & 33.04 & 68.13 & 63.72 \\
 \hline
\end{tabular}
\end{center}
\caption{Similarity evaluation results($\rho \times 100$) on small corpus. Sym-999 denotes Symlex-999 dataset.}
\label{table:full word similarity small corpus}
\end{table}

\begin{table}[!ht]
\begin{center}
\begin{tabular}{ |r|r|r|r| }
 \hline
 & Sim-999 & WS-353 & MEN-3k \\
 \hline
 CBOW & 37.56 & 63.46 & 70.53 \\
 \hline
 SG & 36.88 & 71.65 & 74.81 \\
 \hline
 DSG & 38.32 & 70.07 & 73.38 \\
 \hline
 SSSG & 37.75 & 70.66 & 73.83 \\
 \hline
 FastText & 37.36 & 73.53 & 76.43 \\
 \hline
 \hline
 $CSG_{EF,0.25}$ & 36.67 & 71.64 & 74.69 \\
 \hline
 $CSG_{LF,0.25}$ & 36.94 & 71.54 & 74.78 \\
 \hline
 $CSG_{EF,0.5}$ & 40.38 & 72.81 & 75.62 \\
 \hline
 $CSG_{LF,0.5}$ & 38.59 & 72.45 & 76.71 \\
 \hline
 $CSG_{EF,0.75}$ & 41.29 & 72.87 & 75.88 \\
 \hline
 $CSG_{LF,0.75}$ & 39.97 & \textbf{73.16} & \textbf{76.95} \\
 \hline
 $CSG_{EF,1}$ & \textbf{42.51} & 72.40 & 75.81 \\
 \hline
 $CSG_{EF,0 \to 1}$ & 41.54 & 72.66 & 76.22 \\
 \hline
 $CSG_{LF,0 \to 1}$ & 41.39 & 72.62 & 76.53 \\
 \hline
 $CSG_{EF,ran}$ & 39.60 & 72.46 & 75.85 \\
 \hline
 $CSG_{LF,ran}$ & 38.52 & 72.82 & 76.56 \\
 \hline
\end{tabular}
\end{center}
\caption{Similarity evaluation results($\rho \times 100$) on the large corpus. Sym-999 denotes Symlex-999 dataset.}
\label{table:full word similarity large corpus}
\end{table}

Our results are reported in Tables \ref{table:full word similarity small corpus} and \ref{table:full word similarity large corpus}. Word similarity evaluation results on different corpora are in the Appendix. The CBOW model has degraded performance on the small corpus because it has less effective training samples than the SG based models. The FastText model yielded comparable overall results with the CSG models on this task, especially in the MEN-3k dataset.

On both corpora, the CSG models show superior overall performance than the baseline models. The performance effects of the fusion method depend on the corpus size and pre-processing style. For static fusion weights, LF works better on the small corpus, while EF works better on the large corpus. Furthermore, the results show that a higher $\gamma$ leads to better scores. However, when $\gamma$ becomes one, WS-353 and MEN-3k scores dropped on both corpora. This is considered to happen due to the absence of a direct relationship between the center word and surrounding word during training. The linear fusion weight scheme, $\gamma_{linear, 0 \to 1}$, helps to settle this issue. As a result, $\gamma_{linear, 0 \to 1}$ achieved a balanced and superior overall score. The $\gamma_{ran}$ shows similar result to $\gamma=0.5$ but seems less dependent on context information than $\gamma=0.5$.

\begin{table}[!ht]
\begin{center}
\begin{tabular}{ |r|r r r| }
 \hline
  & \multicolumn{2}{c}{Google} & MSR \\
  & Semantic & Syntactic & Syntactic \\
 \hline
 CBOW & 57.95 & 64.57 & 52.55 \\
 \hline
 SG & 54.54 & 59.58 & 46.68 \\
 \hline
 DSG & 55.74 & 63.02 & 49.67 \\
 \hline
 SSSG & 56.38 & 62.56 & 49.15 \\
 \hline
 FastText & 40.24 & 55.13 & 43.22 \\
 \hline
 \hline
 $CSG_{EF,0.25}$ & 54.23 & 59.20 & 46.12 \\
 \hline
 $CSG_{LF,0.25}$ & 54.53 & 59.28 & 46.04 \\
 \hline
 $CSG_{EF,0.5}$ & 58.91 & 64.53 & 50.30 \\
 \hline
 $CSG_{LF,0.5}$ & 56.92 & 62.23 & 48.85 \\
 \hline
 $CSG_{EF,0.75}$ & 59.86 & 65.77 & 51.56 \\
 \hline
 $CSG_{LF,0.75}$ & 58.57 & 63.76 & 49.81 \\
 \hline
 $CSG_{EF,1}$ & 59.44 & \textbf{66.32} & 52.40 \\
 \hline
 $CSG_{EF,0 \to 1}$ & \textbf{60.05} & 65.35 & \textbf{52.63} \\
 \hline
 $CSG_{LF,0 \to 1}$ & 59.10 & 66.05 & 51.51 \\
 \hline
 $CSG_{EF,ran}$ & 58.23 & 63.62 & 49.46 \\
 \hline
 $CSG_{LF,ran}$ & 56.90 & 62.26 & 48.72 \\
 \hline
\end{tabular}
\end{center}
\caption{Top 1 accuracy for word analogy tasks on large corpus.}
\label{table:full word analogy}
\end{table}

\subsection{Word Analogy Task}
The word analogy task is to solve questions that require predicting a word D from given words A, B, C, and the relationship "A is to B as C is to D". The analogy task is solved by finding a word vector most similar to B + C - A. The accuracy of each model is measured by counting the correct answers. We employed two different datasets for this task: the Google analogy dataset \cite{41224} and the MSR analogy dataset \cite{Mikolov-etal-2013-linguistic}. The Google dataset involves 10,675 syntactic and 8,869 semantic questions, while the MSR dataset is composed of 8,000 syntactic questions. The large corpus was used to minimize the number of unanswerable questions.

Table \ref{table:full word analogy} presents results on the word analogy task. As word similarity task, the CSG models trained with high $\gamma$ give decent results. Accordingly, $\gamma_{linear, 0 \to 1}$ achieves the best overall result in this task. Interestingly, the CBOW shows the best performance among the baselines in contrast with the word similarity evaluation task. Hence, leveraging context information seems to improve performance in this task.

\begin{table}[!h]
\begin{center}
\begin{tabular}{ r|r r r r r r }
 \hline
  & dev F1 & test F1 \\
 \hline
 CBOW & 93.21 & 89.52 \\
 \hline
 SG & 94.22 & 90.22 \\
 \hline
 DSG & \textbf{94.48} & 90.82 \\
 \hline
 SSSG & 94.27 & \textbf{90.89} \\
 \hline
 FastText & 94.43 & 90.55 \\
 \hline
 \hline
 $CSG_{EF,0.25}$ & 94.22 & 90.39 \\
 \hline
 $CSG_{LF,0.25}$ & 94.16 & 90.27 \\
 \hline
 $CSG_{EF,0.5}$ & 94.02 & 89.76 \\
 \hline
 $CSG_{LF,0.5}$ & 94.12 & 90.12 \\
 \hline
 $CSG_{EF,0.75}$ & 94.14 & 90.00 \\
 \hline
 $CSG_{LF,0.75}$ & 94.09 & 90.22 \\
 \hline
 $CSG_{EF,1}$ & 94.07 & 89.90 \\
 \hline
 $CSG_{EF,0 \to 1}$ & 94.04 & 90.11 \\
 \hline
 $CSG_{LF,0 \to 1}$ & 94.18 & 90.12 \\
 \hline
 $CSG_{EF, ran}$ & 94.18 & 90.31 \\
 \hline
 $CSG_{LF, ran}$ & 94.22 & 90.48 \\
 \hline
\end{tabular}
\end{center}
\caption{F1 score of CoNLL 2003 dev/train set.}
\label{table:full ner result}
\end{table}

\subsection{Named Entity Recognition}
For extrinsic evaluation, we conducted a named entity recognition(NER) task. A CoNLL-2003 English \cite{tjong-kim-sang-de-meulder-2003-introduction} benchmark dataset, containing train/dev/test sets, was used. A bidirectional LSTM-CRF model \cite{DBLP:journals/corr/HuangXY15, lample-etal-2016-neural} initialized with the produced embeddings was used to make predictions. During training, the parameter set with the best dev set F1 score was selected as the output.

Table \ref{table:full ner result} shows the results. Contrary to previous tasks, the CSG models trained with high $\gamma$ achieved similar or degraded performance on the NER task. On the other hand, low context weights and $\gamma_{ran}$ present similar or better F1 scores than the original SG model. Still, other SG augmentations are better than $\gamma_{ran}$ on the NER task. From the poor result of the CBOW, we guess that utilizing context information does not fit well with the NER task.

\section{Conclusion}
In this paper, we presented a simple but strong augmentation that utilizes both the context information and the center word without extra out of corpus resources. The experiment results show that the CSG model could provide finer pre-trained word representations. In addition, the CSG model could potentially achieve better performance with further research on sophisticated context function and fusion weight scheme.

\bibliographystyle{IEEE-trans}
\bibliography{references} 

\appendix
\section{Experiments}
\subsection{Word Similarity Evaluation}
Table \ref{table:text8 word similarity} shows full results of \textit{text8} corpus which has different pre-porcessing style and much smaller vocabulary size.\footnote{http://mattmahoney.net/dc/textdata.html} From the results of static fusion weights, trade-off between scores of datasets is observed when fusion weights go high. The CSG models trained with $\gamma_{linear, 0 \to 1}$ also present good performance on this corpus.

\begin{table}[!ht]
\begin{center}
\begin{tabular}{ |r|r|r|r| }
 \hline
 & Sim-999 & WS-353 & MEN-3k \\
 \hline
 CBOW & 30.62 & 70.83 & 59.28 \\
 \hline
 SG & 33.11 & 68.57 & 59.43 \\
 \hline
 DSG & 30.02 & 69.18 & 66.95 \\
 \hline
 SSSG & 32.60 & 66.77 & 56.99 \\
 \hline
 FastText & 31.21 & 64.37 & 59.91 \\
 \hline
 \hline
 $CSG_{EF,0.25}$ & 33.65 & 70.22 & 60.02 \\
 \hline
 $CSG_{LF,0.25}$ & 33.23 & 69.51 & \textbf{60.31} \\
 \hline
 $CSG_{EF,0.5}$ & 33.68 & 69.94 & 60.08 \\
 \hline
 $CSG_{LF,0.5}$ & 33.39 & 70.17 & 60.27 \\
 \hline
 $CSG_{EF,0.75}$ & 33.39 & 70.49 & 60.02 \\
 \hline
 $CSG_{LF,0.75}$ & 33.49 & 70.20 & 59.99 \\
 \hline
 $CSG_{EF,1}$ & 32.90 & 69.97 & 59.50 \\
 \hline
 $CSG_{EF,0 \to 1}$ & \textbf{34.18} & 70.76 & 58.75 \\
 \hline
 $CSG_{LF,0 \to 1}$ & 33.51 & 69.85 & 58.36 \\
 \hline
 $CSG_{EF,ran}$ & 32.76 & \textbf{70.78} & 60.07 \\
 \hline
 $CSG_{LF,ran}$ & 32.98 & 70.39 & 60.07 \\
 \hline
\end{tabular}
\end{center}
\caption{Similarity evaluation results($\rho \times 100$) on the \textit{text8} corpus. Sym-999 denotes Symlex-999 dataset.}
\label{table:text8 word similarity}
\end{table}

\end{document}